\documentclass[journal]{IEEEtran}

\usepackage{xcolor,soul,framed} 

\colorlet{shadecolor}{yellow}
\usepackage{graphicx}
\graphicspath{{../pdf/}{../jpeg/}}
\DeclareGraphicsExtensions{.pdf,.jpeg,.png}

\usepackage[cmex10]{amsmath}
\usepackage{amssymb}
\usepackage{array}
\usepackage{mdwmath}
\usepackage{mdwtab}
\usepackage{eqparbox}
\usepackage{url}
\usepackage{graphicx}
\usepackage{amsmath}
\usepackage{caption,subcaption}
\usepackage{hyperref}

\usepackage{algorithm} 
\usepackage{algpseudocode} 
\usepackage{algorithmicx}
\algrenewcommand\algorithmiccomment[1]{\hfill\(\triangleright\) #1}

\usepackage{array}
\usepackage{footnote}
\makesavenoteenv{tabular}
\usepackage{float}
\usepackage{subfloat}
\usepackage{multirow}
\usepackage{booktabs}
\usepackage{makecell}

\begin{document}
\bstctlcite{IEEEexample:BSTcontrol}
    \title{Self-Aware Active Learning Enables Continual Improvement in Autonomous Driving}
  \author{Dong Hu$^{1}$, Chao Huang$^{2}$~\IEEEmembership{Senior Member,~IEEE}, Carman K.M. Lee$^{1}$,~\IEEEmembership{Senior Member,~IEEE,} \\
  Dimitrios Kanoulas$^{3}$,~\IEEEmembership{Senior Member,~IEEE,}
\thanks{Corresponding author: Chao Huang.}
\thanks{$^{1}$Dong Hu and Carman K.M. Lee are with the Department of Industrial and Systems Engineering, The Hong Kong Polytechnic University, Hong Kong (E-mail: dong24.hu@connect.polyu.hk; ckm.lee@polyu.edu.hk).}
\thanks{$^{2}$Chao Huang is with the School of Electrical and Mechanical Engineering, Adelaide University, Australia (E-mail: chao.huang@adelaide.edu.au).}
\thanks{$^{3}$Dimitrios Kanoulas is with the Department of Computer Science, University College London, United Kingdom (E-mail: d.kanoulas@ucl.ac.uk).}

}  

\markboth{\textit{ }}{xx \MakeLowercase{\textit{et al.}}: xx}
\maketitle

\begin{abstract}
Learning-based autonomous driving (AD) systems can perform reliably in familiar conditions, yet rare distribution shifts and long-tail events remain a major source of abrupt failure. A central limitation is that most agents learn primarily from passive experience and lack mechanisms to estimate when their competence is insufficient, seek timely assistance, and convert safety-critical encounters into targeted improvement. Here we present self-aware guided exploration (SAGE), an active learning framework for post-training adaptation in AD. SAGE learns a predictive world model that generates two online intrinsic signals: fear, which estimates short-horizon predictive risk and model uncertainty, and curiosity, which measures novelty through prediction error. Curiosity adaptively calibrates the intervention threshold for fear, allowing the agent to regulate risk in a context-dependent manner. When predicted fear exceeds this adaptive threshold, the agent transfers control to an expert or fallback policy and uses the resulting takeover trajectories for focused imitation learning. In parallel, fear is integrated into policy optimization and evaluation as a safety-oriented constraint to reduce performance regressions during adaptation. We evaluate SAGE in simulated route-transfer tasks, Waymo-based logged driving scenarios, CARLA occlusion hazards, and real-world mobile robot navigation tests. Across these settings, SAGE improves robustness in novel and safety-critical scenarios, reduces safety violations, and maintains task performance comparable to strong baseline policies. These results suggest that agents can improve after initial training by estimating the limits of their competence, requesting guidance when needed, and learning selectively from rare high-value events.
\end{abstract}

\IEEEpeerreviewmaketitle

\section{Introduction}

\IEEEPARstart{L}{earning}-based autonomous driving (AD) systems have made substantial progress in perception, prediction, and closed-loop control, yet their reliability under rare and safety-critical distribution shifts remains limited \cite{geisslinger2023ethical}. Most learning-based driving pipelines remain largely passive: data are collected or curated offline, failures are identified after the fact, and policies are periodically retrained rather than updated through selective, safety-supervised interaction. Although such pipelines can achieve strong performance in routine conditions, they provide limited mechanisms for agents to estimate the boundaries of their competence, seek assistance in unfamiliar or high-risk situations, or incorporate targeted feedback under safety supervision \cite{yu2024online, chen2025predicting}. In open-world driving, however, such situations may arise after training, including unusual road geometries, unexpected human behaviors, extreme weather and complex multi-agent interactions \cite{feng2023dense}.

Reinforcement learning (RL) provides a general framework for sequential decision-making under uncertainty and has shown strong performance in simulated driving environments \cite{chen2024end}. However, scaling experience alone does not guarantee reliable improvements in safety-critical systems \cite{feng2026breaking}. Two structural limitations remain. First, most learned policies lack explicit awareness of what they do not know: they cannot reliably detect when a situation exceeds their competence or when additional supervision is needed \cite{cao2023continuous}. Second, training and deployment are usually separated. After training, policies are often treated as static artifacts, updated only through offline retraining cycles, which limits their ability to adapt safely to new conditions \cite{dohare2024loss}. As a result, simply accumulating more data does not necessarily raise the safety floor and may even introduce regressions. Recent efforts address these problems through larger datasets \cite{yan2023learning, wang2025data}, domain randomization \cite{wu2023human}, safety filters \cite{pek2020using} and constrained optimization \cite{he2023fear}. Although valuable, these approaches remain largely passive: risk reduction is achieved through externally specified objectives, curated datasets or fixed constraints, rather than through an agent that can monitor its own limitations and acquire supervision when needed \cite{kaufmann2023champion, ma2024efficient}.

\begin{figure*}
    \begin{center}
    \includegraphics[width=0.85\linewidth]{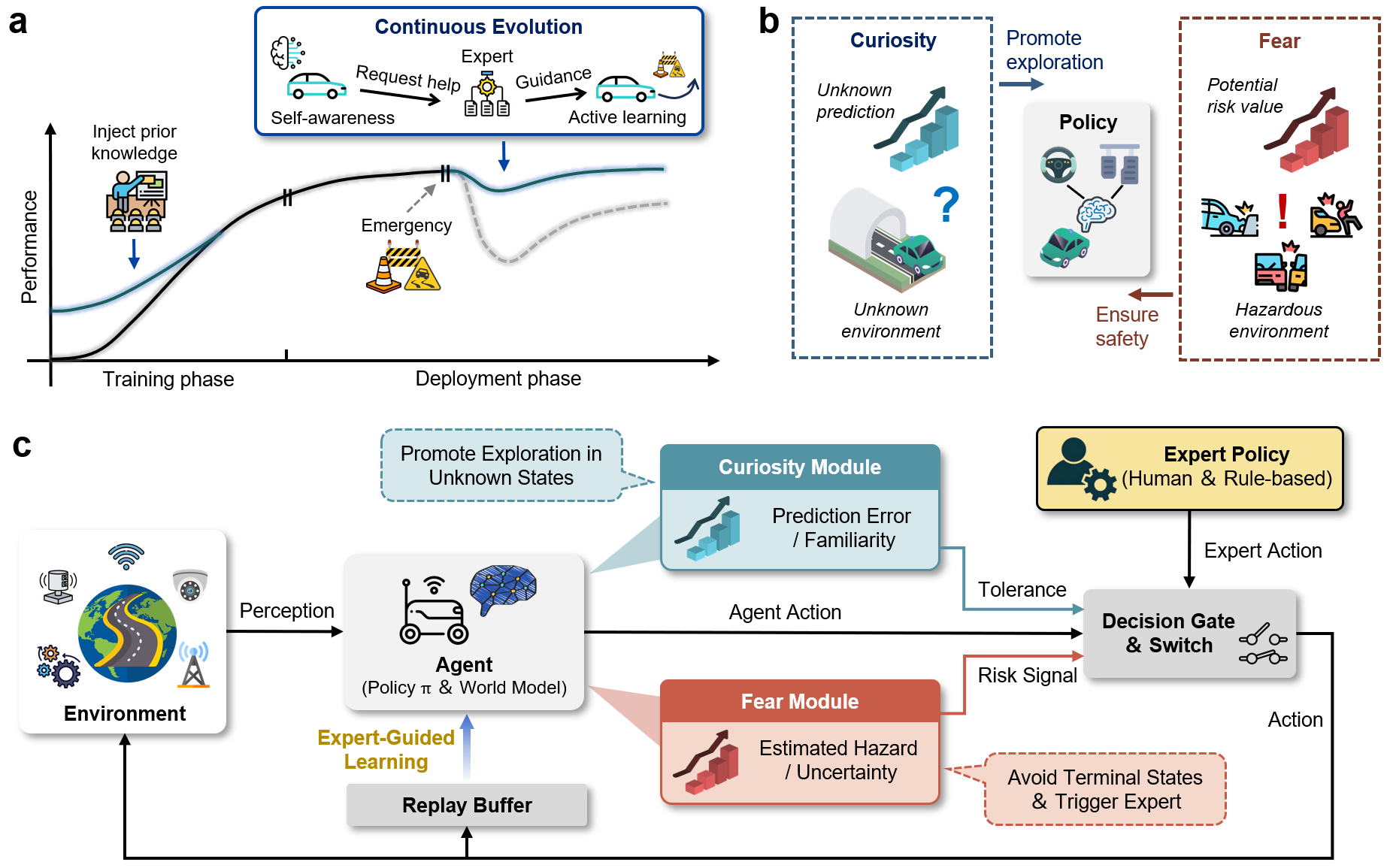}
    \caption{Overview of SAGE for self-aware active learning in autonomous driving.
    (a) A deployed policy may encounter rare distribution shifts that degrade performance. Expert takeover provides targeted corrective data. 
    (b) SAGE estimates novelty and predictive risk from internal signals.
    (c) The framework couples autonomous execution, selective expert intervention, and continual policy adaptation.} 
    \label{framework_1}
 \end{center}                                       
\end{figure*}

In practice, deployed AD systems often rely on highly engineered autonomy stacks and fallback controllers, which can handle many routine scenarios reliably \cite{schwall2020waymo}. The residual challenge is therefore not ordinary driving, but identifying rare situations in which performance may degrade, triggering targeted assistance, and learning from those interventions without compromising safety. Here, we propose self-aware guided exploration (SAGE), an active learning framework for safe continual adaptation after training. SAGE equips the agent with internal signals that estimate both how unfamiliar the current situation is and how hazardous it may become, and uses these signals to decide when autonomous execution is reliable, when expert assistance is needed, and how intervention data should be converted into targeted policy improvement. As illustrated in Fig.~\ref{framework_1}, SAGE combines a learned world model, intrinsic estimates of curiosity and fear, a selective expert intervention mechanism, and continual policy updates from takeover trajectories. Building on a competent baseline policy, SAGE closes the loop between self-monitoring, active learning and safety-preserving policy improvement after training, enabling targeted adaptation to rare or unexpected events (Fig.~\ref{framework_1}(a)).

We use self-awareness in an operational sense: SAGE estimates novelty, predictive risk, and model uncertainty through two complementary intrinsic signals inspired by biological learning: curiosity, which captures novelty through prediction error and highlights poorly modeled regions \cite{pathak2017curiosity}, and fear, which aggregates predicted safety cost and epistemic uncertainty over a short horizon \cite{ledoux2018surviving, averbeck2017motivational}. These signals are derived from a learned world model that predicts the next ego-centric state and safety-relevant cost conditioned on the current state-action pair (Fig.~\ref{framework_1}(c)). A decision gate combines curiosity and fear to arbitrate control between the learned policy and an expert policy, which may be human- or rule-based. Curiosity modulates how much predicted danger the agent is willing to tolerate, so that intervention is triggered only when predicted fear exceeds a curiosity-dependent threshold. The resulting takeover trajectories are stored as targeted supervision for few-shot imitation and continual updates, while fear is also incorporated into policy optimization as a safety-oriented regularizer to reduce regressions.

The main contributions of this work are fourfold:
\begin{enumerate}
    \item A self-aware active learning framework for continual driving adaptation. We propose SAGE, which couples predictive self-monitoring, selective expert intervention, and policy improvement after deployment.
    \item A fear-curiosity intervention mechanism. We define fear as a short-horizon predictive risk signal derived from expected safety cost and epistemic uncertainty, and use curiosity to adaptively calibrate the intervention threshold.
    \item A safety-regularized learning objective with targeted imitation. Intervention trajectories are incorporated through value-weighted imitation, while fear regularization constrains policy updates to reduce unsafe exploration and regressions.
    \item Comprehensive evaluation across simulation, logged driving scenarios, long-tail hazards, and real-world field tests. Experiments demonstrate improved robustness and safety under route shifts, adversarial traffic interactions, occluded pedestrian events, and physical unmanned ground vehicle (UGV) navigation.
\end{enumerate}

Our approach differs from established paradigms in safe and data-efficient RL. Existing methods typically address only one part of the deployment problem. Safe policy improvement, constrained exploration, and related formulations reduce risk by restricting updates or biasing behavior toward conservative actions \cite{yu2022reachability, thomas2019preventing, cao2023continuous}, but often require substantial interaction data and may become unreliable under severe distribution shift. Constraint-based methods and runtime safety filters impose predefined limits or override unsafe actions \cite{garcia2015comprehensive,berkenkamp2017safe, alshiekh2018safe}, yet the diversity of open-world driving is difficult to fully capture with a fixed set of constraints, rules, or models. Offline RL mitigates biases from unsupported actions in static datasets \cite{levine2020offline, wang2025data}, and uncertainty estimation or out-of-distribution detection can flag anomalous or uncertain states \cite{geisslinger2023ethical, wang2025uncertainty}, but neither provides a mechanism to defer control, acquire targeted supervision, and adapt safely after deployment. Approaches aimed at rapid adaptation, such as meta-learning and lifelong RL, seek to improve transfer and continual learning efficiency \cite{o2022neural, meng2025preserving}, but do not provide a mechanism by themselves to decide when adaptation is needed or to ensure that adaptation proceeds safely in high-risk settings. Imitation-based correction methods, such as DAgger \cite{ross2011reduction}, use expert feedback to mitigate compounding errors. More broadly, human-in-the-loop paradigms extend this reliance on human supervision to deployment-time monitoring and correction, making adaptation labor-intensive, difficult to scale, and hard to standardize \cite{wu2023toward,wu2023human}. Overall, prior paradigms lack an integrated framework that can recognize the limits of competence, request targeted expert intervention when necessary, and convert those interventions into safety-regularized continual learning.

Together, SAGE provides a unified framework for adapting to distribution shifts encountered after training. Its design can be summarized by three interacting elements:
\begin{enumerate}
    \item World model-based self-awareness: a predictive dynamics model estimates future outcomes and uncertainty, producing real-time novelty and fear signals.
    \item Selective expert intervention: a decision rule uses these signals to determine whether control remains with the learned policy or transfers to the expert policy, while logging takeover segments as targeted supervision.
    \item Continual constrained adaptation: the policy is updated from intervention trajectories through imitation-guided RL with safety-oriented regularization.
\end{enumerate}

We validate SAGE in four complementary settings: Experiment 1 studies long-term goal-driven navigation, Experiment 2 evaluates policy behavior on Waymo-based logged driving scenarios, Experiment 3 focuses on safety-critical long-tail scenarios, and Experiment 4 presents real-world field tests. Across routine and safety-critical scenarios, SAGE supports adaptive behavior under unexpected events, invokes targeted intervention when needed, and improves safety-related metrics while preserving task completion.

\begin{figure*}
    \begin{center}
    \includegraphics[width=0.95\linewidth]{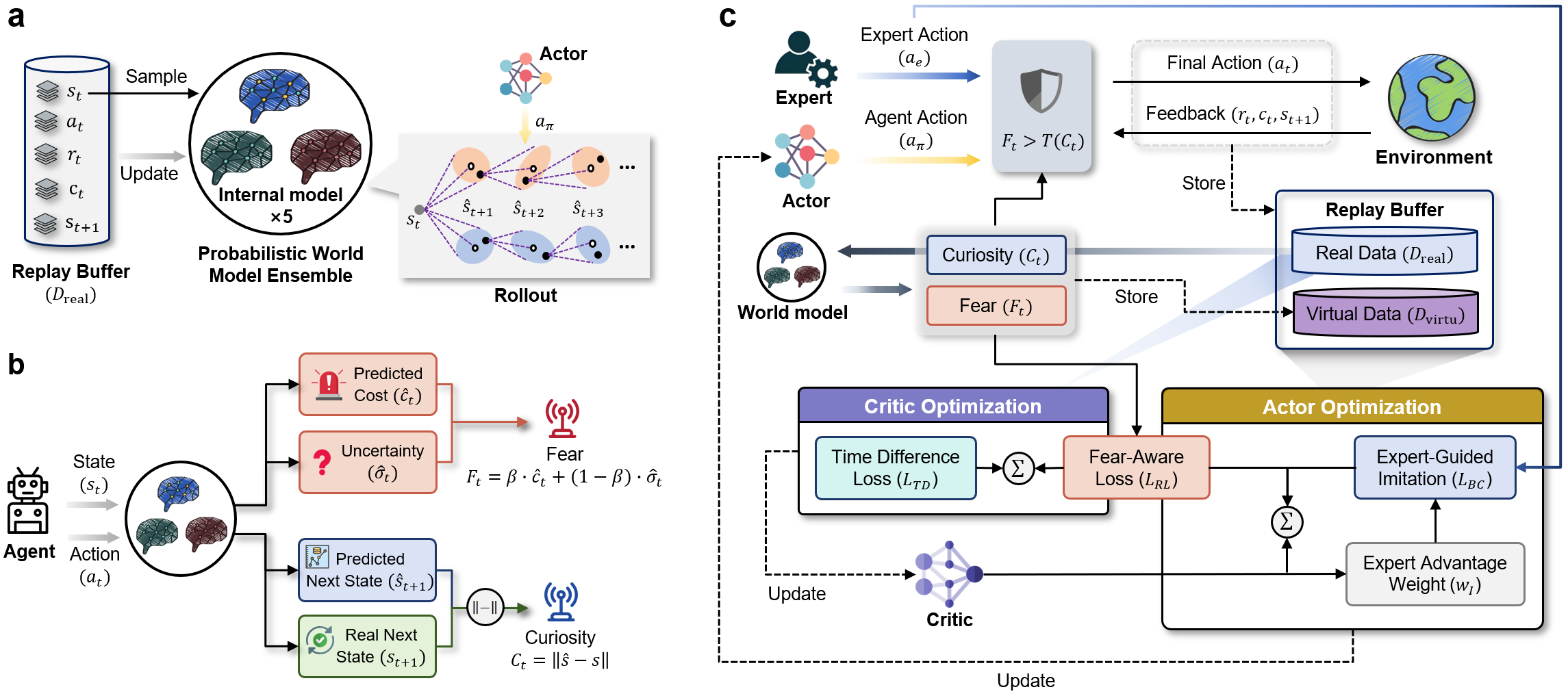}
    \caption{Fear-curiosity gating and policy update mechanism.
    (a) An ensemble world model predicts short-horizon outcomes.
    (b) Prediction error defines curiosity, while expected safety cost and uncertainty define fear. 
    (c) When fear exceeds the curiosity-modulated threshold, expert takeover is activated and the resulting transitions are used for fear-regularized RL and value-weighted imitation.} 
\label{framework_2}
 \end{center}                                       
\end{figure*}

\section{Methods}\label{Sec_Method}

\subsection{Overview}

Fig.~\ref{framework_2} summarizes the implementation of this self-aware active learning mechanism. The core idea is to equip the agent with an internal predictive process that evaluates both the reliability of its intended action and the local knowledge gap. Rather than imposing safety as a fixed external constraint, SAGE balances caution and exploration through three coupled components: prospective fear, curiosity-regulated fear tolerance, and reflex-like expert intervention. At each time step, the agent uses the world model to imagine short-horizon futures from the current observation and candidate action (Fig.~\ref{framework_2}(a)). Predicted costs and model uncertainty are combined into a scalar fear signal, whereas curiosity is quantified by prediction error and reflects how poorly the current situation is modeled (Fig.~\ref{framework_2}(b)). Curiosity then calibrates an adaptive safety threshold \(\tau(C_t)\). When predicted fear exceeds this curiosity-regulated tolerance, i.e., \(F_t > \tau(C_t)\), control is temporarily transferred to an expert controller, implemented as a human operator, rule-based controller, or separately trained policy depending on the experiment (Fig.~\ref{framework_2}(c)), analogous to a reflex overriding habitual behavior under dangerous or poorly understood conditions. Importantly, intervention is not only protective but also instructional: takeover segments are logged in the replay buffer and reused as targeted supervision for policy improvement.

\subsection{Unified Safety-constrained Framework}

We formulate the task as a constrained Markov decision process (CMDP) with an explicit safety cost. Formally, a CMDP is defined by the tuple $(\mathcal{S}, \mathcal{A}, p, r, c, \gamma)$, where $\mathcal{S}$ is the state space, $\mathcal{A}$ is the action space, $p$ denotes the transition dynamics, $r$ is the reward function, $c$ is a non-negative safety cost function, and $\gamma \in (0,1)$ is the discount factor. The standard CMDP objective is to maximize the expected cumulative reward while constraining the expected cumulative safety cost:
\begin{equation}
\max_{\pi} \; \mathbb{E}_{\pi} \left[ \sum_{t=0}^{\infty} \gamma^t r(s_t, a_t) \right]
\quad \text{s.t.} \quad
\mathbb{E}_{\pi} \left[ \sum_{t=0}^{\infty} \gamma^t c(s_t, a_t) \right] \le d,
\end{equation}
where $d$ is a predefined cost budget.

\subsection{Prospective Fear from World Model}

Inspired by anticipatory defensive responses, we define fear as a predictive risk signal that combines expected safety cost and epistemic uncertainty. To instantiate this signal, we employ an ensemble-based world model \(\mathcal{M}\) that predicts safety-relevant quantities from a state-action pair \((s_t, a_t)\). Specifically,
\begin{equation}
(\hat{s}_{t+1}, \hat{c}_t) \sim \mathcal{M}(s_t, a_t),
\end{equation}
where $\hat{s}_{t+1}$ is the predicted next state and $\hat{c}_t$ is the predicted cost associated with the transition, such as a collision or traffic-rule violation.

The world model is implemented as an ensemble of $N$ probabilistic dynamics models $\{\mathcal{M}_{\phi_n}\}_{n=1}^{N}$, each parameterized as a diagonal Gaussian \cite{chua2018deep}:
\begin{equation}
\mathcal{M}_{\phi_n}(s_{t+1}, c_t \mid s_t, a_t) =
\mathcal{N}\!\left(\mu_{\phi_n}(s_t,a_t), \,
\sigma^2_{\phi_n}(s_t,a_t)\right),
\end{equation}
where $\phi_n$ denotes the parameters of the $n$-th ensemble member. $\mu_{\phi_n}(s_t,a_t)$ and $\sigma_{\phi_n}(s_t,a_t)$ denote the predicted mean and standard deviation of the modeled transition distribution, respectively. Each ensemble member is trained by minimizing the negative log-likelihood over the real replay buffer $\mathcal{D}_\text{actual}$:
\begin{equation}
\mathcal{J}_{\text{w}}(\phi_n)
= -\mathbb{E}_{(s_t,a_t,c_t,s_{t+1}) \sim \mathcal{D}_\text{actual}}
\left[\log \mathcal{M}_{\phi_n}(s_{t+1}, c_t \mid s_t, a_t)\right].
\end{equation}
Predictions are aggregated across the ensemble to obtain the mean next-state and cost estimates, denoted by $\hat{s}_{t+1}$ and $\hat{c}_t$, together with an uncertainty estimate $\hat{\sigma}_t$:
\begin{equation}
(\hat{s}_{t+1},\hat{c}_t) \sim  \frac{1}{N} \sum_{n=1}^{N} \mu_{\phi_n}(s_t, a_t),
\qquad
\hat{\sigma}_t = \frac{1}{N} \sum_{n=1}^{N} \sigma_{\phi_n}(s_t, a_t),
\end{equation}
where \(\hat{\sigma}_t\) serves as a proxy for epistemic uncertainty, with larger values indicating lower predictive confidence.

We then define the fear signal as a weighted combination of predicted cost and model uncertainty:
\begin{equation}
\label{eq:fear}
F_t = \beta \, \hat{c}_t + (1 - \beta) \, \hat{\sigma}_t,
\qquad
\beta \in [0,1],
\end{equation}
where $\beta$ controls the relative contribution of predicted cost and uncertainty. In implementation, $\hat{c}_t$ is the predicted probability of a safety violation, and $\hat{\sigma}_t$ is normalized to a comparable scale.

To capture risks whose consequences unfold only after several steps, we further extend fear estimation over a short predictive horizon of length $m$. Starting from the current state, the world model is rolled out for $m$ steps to generate predicted state-action sequences $(s_t^m, a_t^m)$. The corresponding short-horizon fear estimate is
\begin{equation}
F_t^{m} = \beta \, \hat{c}(s_t^m, a_t^m) + (1 - \beta) \, \hat{\sigma}(s_t^m, a_t^m),
\end{equation}
where $\hat{c}(s_t^m, a_t^m)$ and $\hat{\sigma}(s_t^m, a_t^m)$ denote, respectively, the predicted safety cost and uncertainty along the rolled-out trajectory. These model-generated transitions are stored in a virtual replay buffer $\mathcal{D}_\text{virtual}$, enabling the agent to reason about near-future danger beyond immediate perception.

\subsection{Curiosity-modulated fear tolerance}

Fear alone is insufficient for efficient learning: excessive fear produces overly conservative behavior, whereas insufficient fear exposes the agent to catastrophic failure. To adapt this trade-off, we introduce a curiosity signal that reflects how poorly the current world model explains newly observed transitions. Specifically, curiosity is quantified by the discrepancy between the predicted and observed next state:
\begin{equation}
\label{eq:curiosity}
C_t = \left\| \hat{s}_{t+1} - s_{t+1} \right\|_2,
\end{equation}
where $\hat{s}_{t+1}$ is the world model prediction of the next state and $s_{t+1}$ is the actual observed next state. Larger prediction errors indicate transitions that are less well captured by the current model and therefore correspond to regions of greater unfamiliarity. Rather than using curiosity as an intrinsic reward, we use it to modulate fear tolerance through a time-varying safety threshold:
\begin{equation}
\label{eq:threshold}
\tau_t = \tau_{\max} - \rho(C_t)\left(\tau_{\max} - \tau_{\min}\right),
\end{equation}
where $\tau_{\min}$ and $\tau_{\max}$ denote the minimum and maximum allowable thresholds, and $\rho(\cdot)$ is a normalization function implemented using an exponential moving average of recent curiosity values. Under this design, higher curiosity lowers the threshold $\tau_t$, making intervention more likely in poorly modeled situations, whereas lower curiosity raises the threshold as the environment becomes more familiar.

\subsection{Fear-regularized Reinforcement Learning}

Having defined fear as an internal predictive signal, we next incorporate it into policy optimization. The key idea is that fear should not act as a rigid hard constraint at every time step, but as a soft regulator that biases learning away from actions expected to lead to dangerous or highly uncertain outcomes.

We therefore formulate learning as a constrained optimization problem in which the agent seeks high long-term task performance while maintaining the average fear level within a tolerable range:
\begin{equation}
\label{eq:fear_cmdp}
\begin{aligned}
\max_{\pi} \quad & \mathbb{E}_{\pi}\left[\sum_{t=0}^{\infty} \gamma^t r(s_t, a_t)\right],
& \text{s.t.} \quad & \mathbb{E}_{\pi}\left[F_t\right] \leq F_0 .
\end{aligned}
\end{equation}
Here, $F_t$ is the fear signal defined in Eq.~(\ref{eq:fear}), and $F_0$ specifies a target upper bound on acceptable internal fear. This formulation does not require fear to vanish entirely; rather, it limits sustained exposure to high anticipated danger. To solve this problem, we apply Lagrangian relaxation:
\begin{equation}
\label{eq:fear_lagrangian}
\mathcal{L}(\pi, \lambda) = \mathbb{E}_{\pi}\left[ \sum_{t=0}^{\infty} \gamma^t r(s_t, a_t) - \lambda \left(F_t - F_0\right)\right],
\end{equation}
where $\lambda \geq 0$ is the dual variable associated with the fear constraint.

\subsubsection{Fear-regularized Policy Evaluation}

Policy evaluation estimates the utility of an action when both task reward and anticipated fear are taken into account. Under a fixed policy $\pi$, we define the fear-regularized Bellman operator $\mathcal{T}_F^\pi$ as:
\begin{equation}
(\mathcal{T}_F^\pi Q)(s_t,a_t) = r(s_t,a_t) +
\gamma \mathbb{E}_{s_{t+1} \sim p(\cdot \mid s_t,a_t)}
\left[ V^{\pi}(s_{t+1}) \right],
\end{equation}
where the value function is given by:
\begin{equation}
V^{\pi}(s_t) =
\mathbb{E}_{a_t \sim \pi(\cdot \mid s_t)}
\left[ Q^{\pi}(s_t,a_t) - \lambda F_t \right].
\end{equation}
Here, $\lambda \ge 0$ controls the influence of fear regularization, and $F_t$ denotes the prospective fear signal associated with the current decision. This formulation captures the intuition that even a high-reward action should be assigned a low value if it is expected to lead into a dangerous region.

Assuming that both the reward and fear signal are bounded, i.e., $|r(s,a)| \le \bar r$ and $F_t \in [\underline{f}, \bar{f}]$, the operator $\mathcal{T}_F^\pi$ remains a $\gamma$-contraction under the supremum norm:
\begin{equation}
\|\mathcal{T}_F^\pi Q_1 - \mathcal{T}_F^\pi Q_2\|_\infty \le 
\gamma \|Q_1 - Q_2\|_\infty.
\end{equation}
Therefore, for any fixed policy $\pi$, the fear-regularized action-value function admits a unique fixed point, and iterative policy evaluation remains well defined and stable. Intuitively, the fear term acts as a bounded additive correction to the return and does not alter the fundamental contractive structure of Bellman evaluation.

For stable learning, we maintain an ensemble of critic networks $\{Q_{\phi^z}\}$, indexed by $z$, and use the minimum target critic estimate to reduce optimistic bias. The critic parameters $\phi^z$ are trained by minimizing:
\begin{equation}
\label{eq:critic_loss_fear}
\mathcal{L}_c(\phi^z) =
\mathbb{E}_{(s_t,a_t,r_t,s_{t+1}) \sim \mathcal{D}_{\mathrm{actual}}}
\left[\left\| y - Q_{\phi^z}(s_t,a_t) \right\|_2^2\right],
\end{equation}
where the temporal-difference target is defined as:
\begin{equation}
y =
\begin{cases}
r_t + \gamma \tilde{Q}(s_{t+1},a_{t+1}), & \text{if no safety violation occurs}, \\
- c^*, & \text{otherwise},
\end{cases}
\end{equation}
with
\begin{equation}
\tilde{Q}(s_{t+1}, a_{t+1})
= \min_{z} Q_{\bar{\phi}^z}(s_{t+1}, a_{t+1}) - \lambda F_{t+1}.
\end{equation}
Here, $a_{t+1} \sim \pi(\cdot \mid s_{t+1})$, and $\bar{\phi}^z$ denotes the corresponding target-network parameters, updated by Polyak averaging. When a safety violation is observed, the transition is treated as terminal and assigned a large negative value $-c^*$. This construction assigns terminally unsafe transitions a large negative target, preventing short-term reward from offsetting catastrophic outcomes.

The penalty magnitude $c^*$ is chosen under a finite-horizon safety assumption. Following \cite{thomas2021safe,he2023fear}, we assume that there exists a finite horizon $H$ such that, once the agent enters an unsafe progression, a safety violation becomes inevitable within at most $H$ steps. Under this assumption, we derive a sufficient penalty level that guarantees any trajectory leading to a safety violation has a lower discounted return than one that remains safe.

Specifically, let $\bar r$ and $\underline r$ denote upper and lower bounds on reward, respectively, and let $F_t \in [\underline f,\bar f]$ be bounded. Then an unsafe trajectory has maximal possible discounted return upper-bounded by:
\begin{equation}
\sum_{t=0}^{H-1}\gamma^t \bar r
+ \sum_{t=H}^{\infty}\gamma^t (-c^*)
- \lambda \underline f
= \frac{\bar r(1-\gamma^H)-c^*\gamma^H}{1-\gamma}
- \lambda \underline f,
\end{equation}
whereas a safe trajectory has minimal discounted return lower-bounded by:
\begin{equation}
\sum_{t=0}^{\infty}\gamma^t \underline r
- \lambda \bar f
= \frac{\underline r}{1-\gamma}
- \lambda \bar f.
\end{equation}
To ensure that unsafe trajectories are always evaluated as less desirable than safe ones, it suffices to impose:
\begin{equation}
\frac{\bar r(1-\gamma^H)-c^*\gamma^H}{1-\gamma}
- \lambda \underline f
< \frac{\underline r}{1-\gamma}
- \lambda \bar f.
\end{equation}
Rearranging yields the sufficient condition:
\begin{equation}
c^* > \frac{\bar r - \underline r + \lambda(1-\gamma)(\bar f - \underline f)}{\gamma^H} - \bar r.
\end{equation}
Since the fear signal is bounded, we choose:
\begin{equation}
\label{eq:penalty}
c^* = \frac{\bar r - \underline r}{\gamma^H(1-\gamma)}
+ \frac{\lambda}{\gamma^H}
- \frac{\bar r}{1-\gamma},
\end{equation}
which is sufficient when the normalized fear range satisfies $\bar f - \underline f \le 1$. This choice guarantees that trajectories leading to safety violations receive strictly lower value estimates than trajectories that remain safe, thereby preventing unsafe policies from dominating the learning process.

\subsubsection{Fear-regularized policy improvement}

Policy improvement optimizes the actor for high task return under a target upper bound on the fear signal. Using Lagrangian relaxation, the constrained objective is transformed into the following saddle-point problem:
\begin{equation}
\max_{\pi} \min_{\lambda \ge 0}\;
\mathbb{E}
\left[ \sum_{t=0}^{\infty} \gamma^t r(s_t,a_t)
+ \lambda (F_0 - F_t) \right],
\end{equation}
where $F_0$ is the desired fear threshold and $\lambda$ is the dual variable that enforces the fear constraint.

Under this formulation, policy improvement is performed by maximizing a fear-regularized action value, which yields the actor objective:
\begin{equation}
\label{eq:actor_loss_fear}
\mathcal{L}_{\text{a}}^{\text{RL}}(\theta) =
\mathbb{E}_{s_t \sim \mathcal{D}_\text{actual} \cup \mathcal{D}_\text{virtual}, \;
a_t \sim \pi_{\theta}(\cdot|s_t)}
\left[\lambda F_t - Q(s_t, a_t)\right].
\end{equation}
This objective encourages the policy to prefer high-value actions while avoiding those with excessive predicted fear. Because the fear term is derived from predicted future outcomes rather than immediate unsafe feedback, it enables proactive risk avoidance before violations occur.

The dual variable $\lambda$ is updated to adaptively regulate the overall fear level:
\begin{equation}
\label{eq:lambda_update}
\mathcal{L}_{\lambda} =
\mathbb{E}_{s_t \sim \mathcal{D}_\text{actual}}
\left[\lambda (F_0 - F_t)\right], \qquad \lambda \ge 0.
\end{equation}
Here, $\lambda$ is optimized on the dual side and projected onto the nonnegative domain after each update. If the average fear exceeds the target $F_0$, the dual update increases $\lambda$, thereby amplifying the fear penalty in subsequent actor updates. Conversely, when the fear level remains below the target, $\lambda$ decreases, allowing the policy to behave less conservatively. This adaptive mechanism provides automatic regulation of risk sensitivity during training and removes the need to manually tune a fixed penalty coefficient for all stages of learning.

\subsection{Active Expert Guidance Mechanism}

When predicted fear exceeds the adaptive threshold, autonomous execution is considered unreliable, and control is temporarily transferred to an expert policy. The resulting intervention serves two purposes: at the control level, it reduces unsafe autonomous actions; at the learning level, it provides targeted demonstrations for policy improvement. We implement this process using an active expert intervention framework. Specifically, an intervention indicator $\Xi_t \in \{0,1\}$ is defined as:
\begin{equation}
\Xi_t =
\begin{cases}
1, & \text{if } F_t > \tau_t, \\
0, & \text{otherwise},
\end{cases}
\end{equation}
where $\tau_t$ is the curiosity-regulated threshold. Once fear exceeds this threshold, the agent is deemed to be operating outside its reliable competence envelope. The final action executed in the environment is given by:
\begin{equation}
\label{eq:expert_override}
a_t = (1 - \Xi_t)\, a_t^{\text{RL}} + \Xi_t\, a_t^{\text{expert}},
\end{equation}
where $a_t^{\text{RL}}$ is the action produced by the learned policy and $a_t^{\text{expert}}$ is the action provided by the expert controller. To avoid oscillatory switching caused by transient fear spikes, expert control, once activated, is maintained for a short temporal horizon. This yields temporally coherent corrective behavior rather than unstable frame-by-frame takeover. All transitions are stored in the real replay buffer $\mathcal{D}_{\text{actual}}$. Intervention steps are identified by the indicator $\Xi_t=1$, together with the associated expert action $a_t^{\text{expert}}$ and policy action $a_t^{\text{RL}}$.

These interventions serve not only as an online safety mechanism, but also as a source of targeted supervision for policy learning. Rather than treating all demonstrations as equally informative, we weight expert actions according to their estimated value advantage over the agent's current behavior. The actor is therefore optimized using a composite objective:
\begin{equation}\label{eq:actor_loss_expert}
\mathcal{L}_{\text{a}}(\theta) =
\mathcal{L}_{\text{a}}^{\text{RL}}(\theta) + \mathcal{L}_{\text{BC}}.
\end{equation}
where $\mathcal{L}_{\text{a}}^{\text{RL}}(\theta)$ is the fear-regularized reinforcement objective in Eq.~(\ref{eq:actor_loss_fear}) and $\mathcal{L}_{\text{BC}}$ is a behavior-cloning term applied to expert-labeled states.

For intervention-labeled states in $\mathcal{D}_{\text{actual}}$ with $\Xi_i=1$, we evaluate whether the expert action provides a meaningful improvement over the agent's current behavior by comparing their estimated action values under the learned critic. Rather than using this value difference as a hard decision rule, we convert it into a soft importance weight that continuously modulates the strength of imitation. Concretely, the difference between the Q-values of the expert action and the agent’s action is passed through a smooth and monotonic transformation, so that demonstrations exert a graded influence on policy updates: larger weights are assigned when the expert action is estimated to be superior, whereas smaller weights reduce imitation pressure otherwise. Formally, the importance weight associated with an expert-labeled transition is defined as:
\begin{equation}\label{eq:qa_weight}
\omega_i = \exp\!\left(Q(s_i, a_i^{\text{expert}}) - Q(s_i, a_i^{\text{RL}})
\right),
\end{equation}
which yields a strictly positive and smoothly varying coefficient. When the expert and agent actions have similar estimated values, $\omega_i$ remains close to unity, resulting in only weak imitation pressure. When the expert action exhibits a clear value advantage, the corresponding weight increases, amplifying the contribution of that demonstration to the policy update. Conversely, if the expert action is estimated to be inferior, its influence is naturally reduced.

Through this mechanism, expert demonstrations are selectively incorporated according to their estimated long-term utility rather than blindly imitated. The imitation loss is then defined as:
\begin{equation}\label{eq:bc_loss}
\mathcal{L}_{\text{BC}}
=\mathbb{E}_{(s_i,a_i^{\text{expert}},a_i^{\text{RL}})\sim \mathcal{D}_{\text{actual}},\, \Xi_i=1}
\left[\omega_i \left\| \pi_{\theta}^{\text{RL}}(\cdot \mid s_i) - a_i^{\text{expert}}
\right\|^2 \right].
\end{equation}
where $\pi_{\theta}^{\text{RL}}(\cdot \mid s)$ denotes the action output of the learned stochastic policy conditioned on state $s$. This formulation amplifies expert influence only when the expert action yields higher estimated value than the agent’s own choice, enabling selective and efficient learning from intervention.

Taken together, expert intervention plays a dual role in the proposed framework. At control time, it acts as a safety mechanism that prevents catastrophic actions when fear signals indicate imminent danger. At learning time, it provides targeted supervision in regions of the state space where autonomous decision-making is unreliable. This dual-use design preserves exploration while ensuring that demonstrations are acquired when needed and weighted according to their estimated utility.

\begin{figure*}
    \begin{center}
    \includegraphics[width=0.94\linewidth]{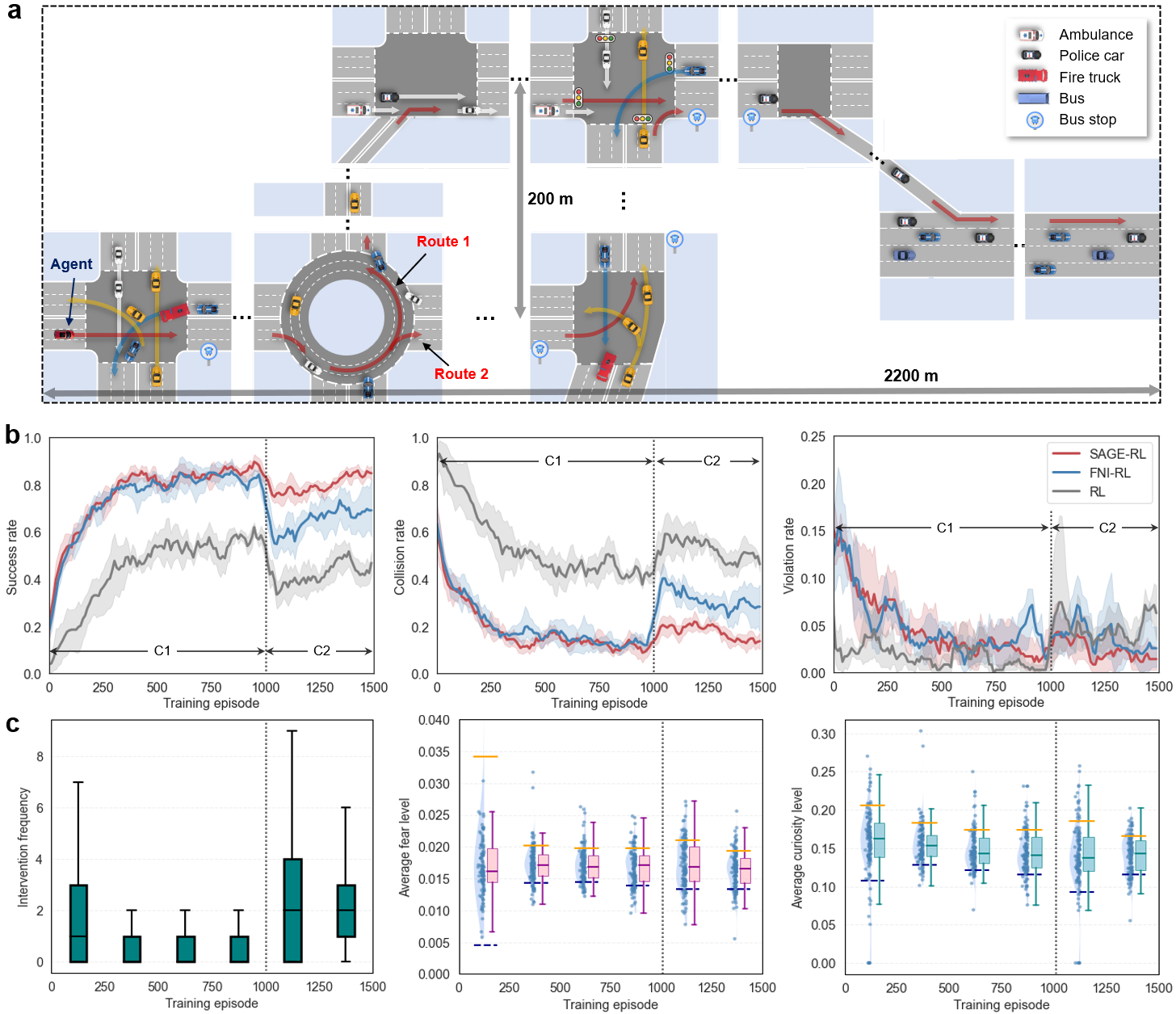}
    \caption{Continual adaptation in a long-term composite traffic-navigation task. 
    (a) Two-phase training protocol on two long-distance routes. Route~1 is used in episodes 1--1000 (C1), after which the agent is switched to Route~2 for episodes 1001--1500 (C2), introducing a distribution shift in road geometry and traffic composition. 
    (b) Training performance of different methods, quantified by success rate, collision rate and red-light violation rate. SAGE-RL shows reduced performance degradation after the route switch and faster recovery in C2. 
    (c) Internal self-awareness signals and intervention statistics of SAGE-RL, including takeover frequency, mean fear and mean curiosity. Fear and intervention frequency increase immediately after the route switch and decline as adaptation progresses.}
    \label{sumo_1}
 \end{center}                                       
\end{figure*}

\begin{figure*}[t]
    \begin{center}
    \includegraphics[width=0.87\linewidth]{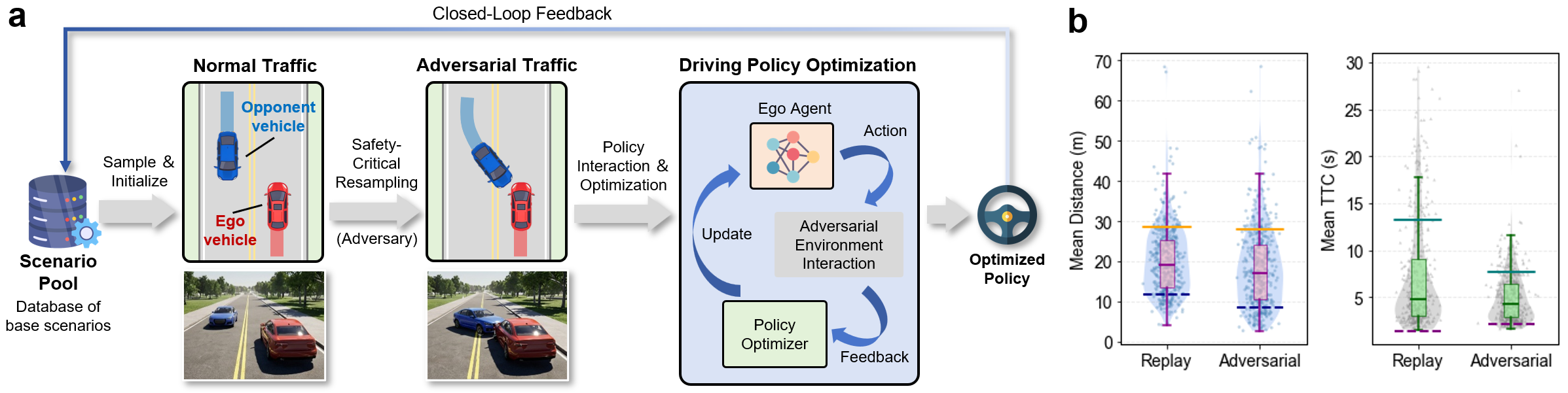}
    \caption{Closed-loop adversarial training on Waymo driving scenarios. 
    (a) Schematic of adversarial scenario generation from logged traffic segments by risk-conditioned resampling of surrounding-agent behaviors. 
    (b) Comparison of opponent-ego interaction statistics in log replay and adversarial scenarios. }
    \label{waymo_1}
 \end{center}                                       
\end{figure*}

\section{Experiments and Results}\label{Sec_Experiments}

\subsection{Baselines}

We compare our method with representative baselines spanning model-free RL, value-based RL, interactive imitation learning, safe model-based RL, offline diffusion-based learning, human-guided RL, and classical navigation planning.

For general RL comparisons, we use SAC~\cite{haarnoja2018soft}, TD3~\cite{fujimoto2018addressing}, and D3QN~\cite{xie2017towards}. SAC and TD3 are standard off-policy actor-critic algorithms for continuous control, while D3QN is adopted in discrete-control settings with double Q-learning and a dueling value architecture. 

To evaluate learning under expert guidance and safety-critical settings, we include HG-DAgger~\cite{kelly2019hg} and Fear-Neuro-Inspired RL (FNI-RL)~\cite{he2023fear}. HG-DAgger represents interactive imitation learning, where expert guidance and demonstration aggregation are used to mitigate distribution shift. FNI-RL is a model-based safe RL method that learns a predictive safety signal inspired by defensive neural mechanisms, providing a strong baseline for proactive risk-aware control.

For the real-world UGV field tests, we further compare with Diffusion Behavior Cloning (Diffusion-BC)~\cite{pearce2023imitating}, Diffusion Q-Learning (Diffusion-QL)~\cite{wang2022diffusion}, Multi-Hug RL~\cite{wu2023human}, and the Timed Elastic Band (TEB) planner~\cite{rosmann2017integrated}. Diffusion-BC and Diffusion-QL are offline diffusion-based methods, where the former learns a conditional diffusion policy from demonstrations and the latter combines diffusion policy modeling with Q-learning. Multi-Hug RL is a human-guided RL framework that leverages interventions, corrective demonstrations, and intervention-based reward shaping for UGV navigation. TEB is a classical optimization-based local planner that generates time-parameterized trajectories under kinodynamic and obstacle-avoidance constraints, serving as a representative traditional navigation baseline.

Unless otherwise specified, all RL experiments use SAC as the backbone. The only exception is the long-tail safety-critical experiments, where D3QN is used due to the discrete action setting. Thus, we use ``RL'' to denote SAC throughout the paper unless explicitly stated.

\begin{figure*}[ht!]
    \begin{center}
    \includegraphics[width=0.9\linewidth]{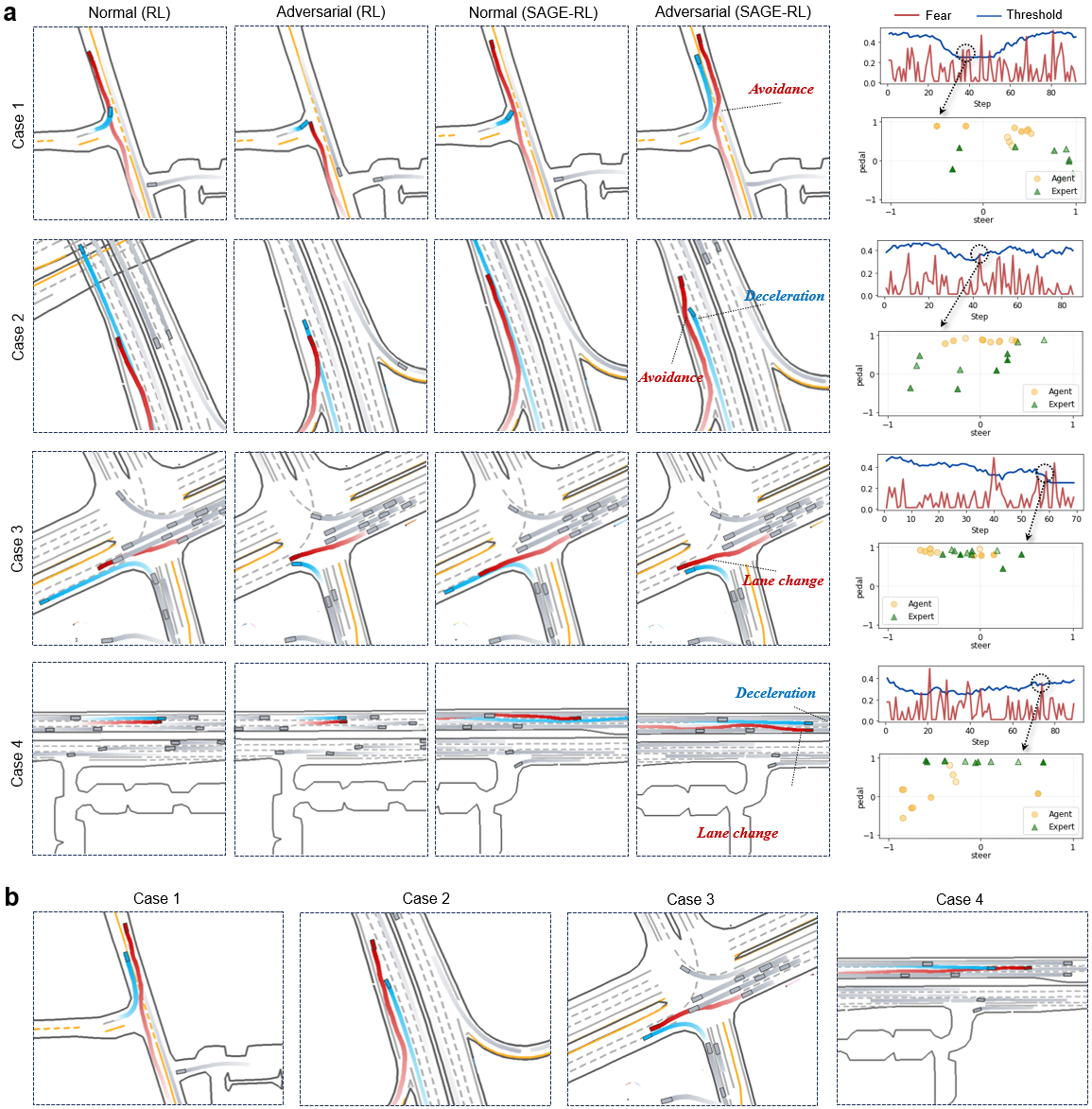}
    \caption{Representative Waymo cases illustrating intervention-triggered safety and learned emergency handling. (a) Four training-time cases comparing a RL baseline and our method under log replay (normal traffic) and adversarial traffic. For each case, the rightmost panels show the fear signal and the curiosity-determined threshold over time (top), together with the agent and expert actions around the intervention window in the raw control space (bottom). (b) Post-training deployment of our method in the corresponding adversarial scenarios without expert intervention.}
\label{waymo_2}
 \end{center}                                       
\end{figure*}

\subsection{Long-Term Goal-Driven Navigation Experiments}

\subsubsection{Task Description}

We evaluate online adaptation to route-level distribution shift on a long-distance composite route (Fig.~\ref{sumo_1}(a)). The route integrates diverse traffic elements, including unsignalized intersections, a roundabout, a signalized intersection, and a highway merge via an on-ramp followed by cruising. Training proceeds in two phases to probe robustness to regime change. In Phase C1 (episodes 1--1000), the agent repeatedly drives along Route~1 until reaching stable performance. In Phase C2 (episodes 1001--1500), the agent is switched to Route~2, which differs in local geometry and traffic composition and introduces unfamiliar scene combinations. To further amplify novelty and interaction complexity, special vehicles (e.g., bus, fire truck, police car, and ambulance) are added during C2. This setting evaluates whether a policy that appears stable in familiar conditions can detect a route-level distribution shift and adapt while limiting performance degradation.

Our method (SAGE-RL) is compared against standard RL and FNI-RL baselines. For SAGE-RL, expert interventions are provided by high-performing standard RL policies trained separately on each route. Specifically, a policy trained on Route~1 supervises Phase C1, while a different policy trained on Route~2 provides supervision during Phase C2. The expert is used only when the intervention gate is activated and is not used for autonomous evaluation.

\subsubsection{Learning Setup}

a) State: At each step, the ego vehicle observes an ego-centric state composed of local traffic context and task cues. We select up to six nearest vehicles within a 200\,m perception range in predefined relative sectors (front, rear, left-front, left-rear, right-front, right-rear). For each neighbor $i\in\mathcal{N}_t$, four relative attributes are encoded: relative distance $d_t^i$, relative heading $\Delta\theta_t^i$, relative speed $\Delta v_t^i$, and relative velocity direction $\Delta\phi_t^i$.  Ego speed $v_t^{\text{ego}}$ and heading $\phi_t^{\text{ego}}$ are appended. If a sector is empty, default values are used. For long-horizon navigation and rule compliance, we further include distance to the next traffic light $d_t^{\text{tl}}$, traffic-light state $\text{tl}_t$, and distance to the next route target $d_t^{\text{nav}}$. The final observation is $s_t\in\mathbb{R}^{29}$.

b) Action: The policy outputs a continuous longitudinal acceleration $a_t\in[a_{\min},a_{\max}]$. Lane changes are handled by the simulator’s built-in routing logic.

c) Reward and Cost: The reward encourages efficiency and goal completion:
\[ r_t = \frac{v_t^{\text{ego}}}{5} + r_t^{\text{goal}} - c_t^{\text{col}} - c_t^{\text{red}}, \]
where $r_t^{\text{goal}}=100$ upon successful off-ramp arrival; otherwise a distance-based shaping term $-\log(1+\Delta d_t/\Delta d_{\max})-1$ is used. Binary safety costs indicate collision $c_t^{\text{col}}\!=\!\mathbb{I}\{\text{collision}\}$ and red-light violation $c_t^{\text{red}}\!=\!\mathbb{I}\{\text{red-light}\}$.

\subsubsection{Results}

Fig.~\ref{sumo_1}(b) reports the success rate, the collision rate, and the red-light violation rate over training. All methods improve steadily during C1, indicating that the composite route is learnable under a fixed distribution. After the route switch at episode 1000, both RL and FNI-RL show a clear drop in success rate together with increased collision and violation rates, suggesting limited robustness to the new route distribution. In contrast, SAGE-RL exhibits markedly lower performance degradation and recovers more quickly in C2, ultimately reaching higher success while maintaining lower collision and violation rates. These results indicate that SAGE-RL supports safer and more effective adaptation after distribution shift.

Fig.~\ref{sumo_1}(c) further relates this performance pattern to the internal signals and intervention dynamics of SAGE-RL. Immediately after the route switch, the average fear level increases, indicating that the world model identifies the new route as unfamiliar and potentially hazardous. This increase is accompanied by a higher takeover frequency, allowing expert intervention to shield the agent from unsafe actions during the transition. As training continues in C2, takeover frequency gradually declines, consistent with the policy improving from intervention data and adapting to the new route. Curiosity remains active throughout this process, providing an additional novelty-related signal for adaptive intervention. Overall, these trends are consistent with a detect-intervene-learn loop that helps stabilize performance under distribution shift and supports continual adaptation to newly introduced rare traffic participants.

\begin{table*}[t]
\centering
\caption{Evaluation results on autonomous driving tasks under normal and adversarial Waymo scenarios.}
\label{tab:test_results}
\renewcommand{\arraystretch}{1.2}
\begin{tabular}{llcccccccc}
\toprule
\multirow{2}{*}{\textbf{Set}}
& \multirow{2}{*}{\textbf{Method}}
& \multicolumn{4}{c}{\textbf{Normal Scenarios}}
& \multicolumn{4}{c}{\textbf{Adversarial Scenarios}} \\
\cmidrule(lr){3-6}
\cmidrule(lr){7-10}
& & \textbf{S.R.(\%)} & \textbf{C.R.(\%)} & \textbf{Speed (m/s)} & \textbf{Acc. (m/s$^2$)}
  & \textbf{S.R.(\%)} & \textbf{C.R.(\%)} & \textbf{Speed (m/s)} & \textbf{Acc. (m/s$^2$)} \\
\midrule
\multirow{4}{*}{Training Set}
& SAC
& 73.9 & 18.9 & $9.42 \pm 3.71$ & $1.34 \pm 0.38$
& 51.7 & 39.2 & $8.64 \pm 4.29$ & $1.76 \pm 0.56$ \\
& TD3
& 75.8 & 18.6 & $9.57 \pm 3.84$ & $1.42 \pm 0.43$
& 48.8 & 42.1 & $8.77 \pm 4.46$ & $1.89 \pm 0.63$ \\
& FNI-RL
& 86.3 & 9.8  & $9.05 \pm 3.62$ & $1.07 \pm 0.29$
& 68.4 & 24.7 & $8.33 \pm 4.08$ & $1.39 \pm 0.42$ \\
& SAGE-RL
& \textbf{91.5} & \textbf{4.8} & $8.96 \pm 3.48$ & \textbf{$0.93 \pm 0.22$}
& \textbf{83.9} & \textbf{9.1} & $8.21 \pm 3.91$ & \textbf{$1.18 \pm 0.33$} \\
\midrule
\multirow{4}{*}{Testing Set}
& SAC
& 71.0 & 21.2 & $9.26 \pm 3.92$ & $1.41 \pm 0.42$
& 46.8 & 43.8 & $8.38 \pm 4.51$ & $1.91 \pm 0.64$ \\
& TD3
& 73.4 & 20.6 & $9.43 \pm 4.03$ & $1.50 \pm 0.47$
& 44.0 & 46.6 & $8.52 \pm 4.67$ & $2.04 \pm 0.72$ \\
& FNI-RL
& 83.8 & 12.4 & $8.87 \pm 3.84$ & $1.15 \pm 0.34$
& 63.8 & 29.0 & $8.07 \pm 4.42$ & $1.52 \pm 0.49$ \\
& SAGE-RL
& \textbf{88.4} & \textbf{8.8} & $8.79 \pm 3.66$ & \textbf{$1.01 \pm 0.27$}
& \textbf{79.6} & \textbf{13.2} & $7.94 \pm 4.21$ & \textbf{$1.29 \pm 0.40$} \\
\bottomrule
&& \\[-1em]
\end{tabular}
\begin{minipage}{0.9\linewidth}
  \footnotesize
  \textit{Note:} S.R. denotes success rate, C.R. denotes collision rate, Speed denotes average driving speed, and Acc. denotes average acceleration.
  Results are averaged over five random seeds. The training and testing sets contain 400 and 100 scenarios, respectively.
\end{minipage}
\end{table*}

\subsection{Waymo-Based Adversarial Driving Experiments}

\subsubsection{Task Description}

We evaluate closed-loop adversarial training using driving scenarios imported from the Waymo Open Motion Dataset \cite{ettinger2021large}. Each base scenario is a short logged segment containing the ego vehicle and surrounding traffic agents. We curate a pool of 500 replay scenarios to initialize training episodes, where surrounding vehicles follow their recorded behaviors.

To expose the policy to rare yet safety-critical interactions, we adopt a closed-loop adversarial training pipeline (Fig.~\ref{waymo_1}(a)) that alternates between generating adversarial variants of logged scenarios and optimizing the ego policy in the resulting environments \cite{zhang2023cat}. For a given base scenario and current policy, opponent trajectories are resampled by reweighting their distribution toward higher-risk outcomes (e.g., sudden cut-ins ahead of the ego vehicle). This induces small-gap and low time-to-collision encounters that require risk-aware responses such as braking and yielding. We verify that adversarial generation indeed shifts the interaction distribution by comparing replay and adversarial settings (Fig.~\ref{waymo_1}(b)): adversarial scenarios exhibit systematically smaller mean inter-vehicle distances and lower time-to-collision, confirming a deliberately elevated-risk training regime.

For SAGE-RL, expert interventions are provided by a separately trained RL policy. This expert policy is trained to convergence using standard RL and serves solely as a supervisory controller when takeover is triggered.

\subsubsection{Learning Setup}
a) State: The observation consists of:
(i) ego kinematics (speed, heading, steering-related states), 
(ii) route guidance (relative distance and direction to the next two checkpoints), and 
(iii) a LiDAR-style encoding with 72 beams and 50\,m range capturing nearby vehicles and road geometry. 
The resulting ego-centric vector is fully observable from onboard sensing.

b) Action: The policy outputs normalized controls $a_t=[a_t^{\text{steer}},a_t^{\text{acc}}]^\top\in[-1,1]^2$. They are mapped to physical steering, acceleration, and braking commands according to simulator limits.

c) Reward and Cost: We adopt the compositional driving reward
\[r_t = (d_t-d_{t-1}) + r_t^{\text{crash}} + r_t^{\text{out}}, \]
where $d_t-d_{t-1}$ measures forward progress, $r_t^{\text{crash}}=-1$ upon collision, and $r_t^{\text{out}}=-10$ when leaving the drivable area. A binary safety cost \[ c_t = \mathbb{I}\{\text{collision} \lor \text{out-of-road}\}\] is used for risk-aware learning.

\begin{figure*}[h!]
    \begin{center}
    \includegraphics[width=0.92\linewidth]{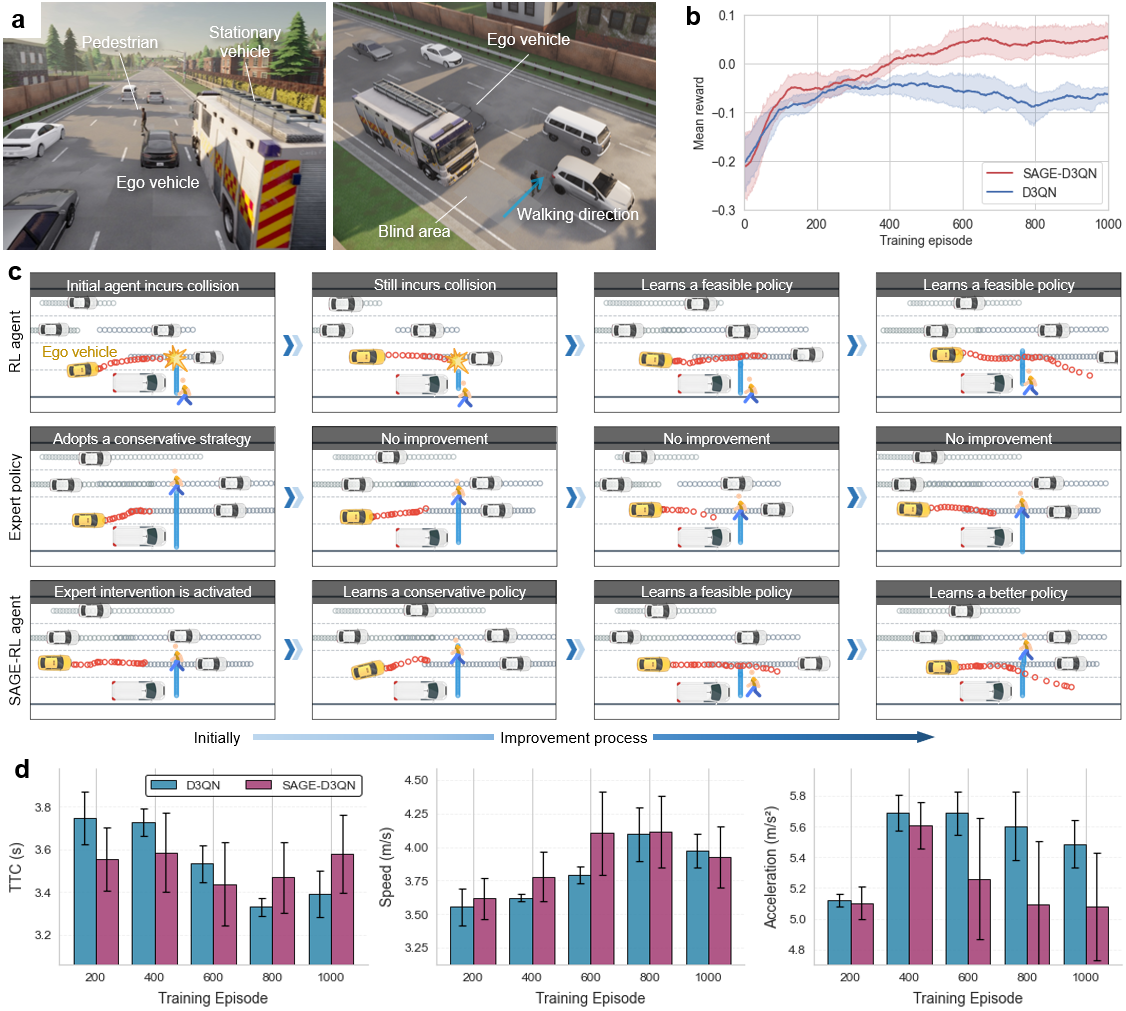}
    \caption{Adaptation to a long-tail safety-critical hazard in CARLA.
    (a) Occlusion-driven pedestrian-crossing scenario in which a parked truck blocks early visibility and the pedestrian enters the ego lane with limited reaction time. 
    (b) Training reward of SAGE-RL and a standard RL baseline. SAGE-RL shows more stable and sustained improvement. 
    (c) Representative multi-agent trajectories at different stages of training, illustrating the transition from early collisions to progressively safer and more calibrated responses in SAGE-RL.
    (d) Quantitative comparison of safety and efficiency metrics across training episodes.}
\label{carla_1}
 \end{center}                                       
\end{figure*}

\subsubsection{Results}

Fig.~\ref{waymo_2}(a) shows four representative training-time cases comparing a standard RL baseline with SAGE-RL in normal and adversarial traffic. In these examples, the RL baseline exhibits unsafe behavior under adversarial interactions, often approaching near-collision states, whereas SAGE-RL produces timely evasive responses. Across all four cases, the fear signal increases sharply as the ego vehicle enters a safety-critical interaction. Once it exceeds the curiosity-regulated intervention threshold, expert takeover is activated. The corresponding action traces indicate that the expert performs scenario-dependent maneuvers rather than simply reducing the control magnitude. Specifically, Case~1 requires an aggressive rightward lane change, Case~2 is resolved by a moderate leftward deviation, and Cases~3-4 favor rightward lane changes, often combined with braking. These interventions avoid collisions in the illustrated cases and provide corrective supervision at critical moments during updating.

After updating, the same scenarios are handled autonomously without expert intervention (Fig.~\ref{waymo_2}(b)). The deployed policy reproduces appropriate emergency responses-timely braking and geometry-consistent lane changes under adversarial traffic, indicating that intervention episodes have been internalized rather than relied on as persistent assistance. Collectively, these examples are consistent with the intended adaptation mechanism: increased risk triggers temporary expert control during training, and the resulting intervention data improve the autonomous handling of similar hazards after updating.

Table~\ref{tab:test_results} compares different RL methods under normal and adversarial Waymo scenarios. Adversarial scenarios consistently degrade policy performance, leading to higher driving risks and less stable control. Compared with SAC, SAGE-RL reduces the collision rate by 76.8\% on the adversarial training set and by 69.9\% on the adversarial testing set, while also producing smoother driving behaviors with lower acceleration. Moreover, SAGE-RL consistently improves the success rate over SAC across both normal and adversarial settings. These results demonstrate that SAGE-RL significantly enhances safety and robustness under challenging driving scenarios without substantially sacrificing driving efficiency.

\subsection{Long-Tail Safety-Critical Experiments}

\subsubsection{Task Description}

We construct a long-tail, safety-critical driving task in CARLA that emphasizes partial observability and reaction under occlusion (Fig.~\ref{carla_1}(a)). The ego vehicle travels along an urban road where a large truck is parked at the roadside. A pedestrian suddenly emerges from in front of the truck and begins crossing the road. Due to the truck’s occlusion, the pedestrian is partially or fully hidden until entering the ego vehicle’s potential path, creating a blind-zone hazard that demands timely braking or evasive action.

This scenario is intentionally rare relative to the agent's prior training distribution: the agent has not previously encountered similar occlusion-induced pedestrian events, and naive exploration often results in immediate collisions. For our method, the expert policy is trained using HG-DAgger, which learns from human driving demonstrations and corrective interventions. This human-guided expert provides supervisory actions during takeover, supplying reliable responses in occlusion-driven pedestrian scenarios.

\subsubsection{Learning Setup}

a) State: A compact ego-centric state is constructed from a 360$^\circ$ LiDAR scan. The azimuth is discretized into 36 bins (10$^\circ$ each), and the nearest return per bin forms $o_t^{\text{lidar}}\in\mathbb{R}^{36}$. We append ego speed $v_t^{\text{ego}}$, distance to goal $d_t^{\text{goal}}$, and goal bearing $\alpha_t^{\text{goal}}$, yielding $s_t=[o_t^{\text{lidar}},v_t^{\text{ego}},d_t^{\text{goal}},\alpha_t^{\text{goal}}]$.

b) Action: A discrete high-level action space with 9 behaviors combines three longitudinal intents (brake, maintain, accelerate) and three lateral intents (left, keep lane, right). Actions are executed every 0.5\,s.

c) Reward and Cost: The shaped reward prioritizes safety and goal achievement:
\[r_t = r_t^{\text{col}} + r_t^{\text{goal}} + 0.1\frac{v_t^{\text{ego}}}{v_{\max}} - 0.1|\delta_t|,\]
where $r_t^{\text{col}}=-10$ upon collision, $r_t^{\text{goal}}=5$ upon goal completion, and $\delta_t$ is the steering magnitude. A binary collision cost $C_t=\mathbb{I}\{\text{collision}\}$ is used as the safety signal.


\subsubsection{Results}

Fig.~\ref{carla_1}(b) shows that SAGE-RL achieves a steadier reward improvement throughout training, while the D3QN baseline plateaus and exhibits pronounced fluctuations. This difference is also visible in the trajectory snapshots in all training stages (Fig.~\ref{carla_1}(c)). In the initial encounters with the occluded pedestrian, the baseline collides. Similar failures remain visible even after extended training, suggesting that repeated exposure to this long-tail hazard is insufficient to produce consistently safe behavior. Similarly, temporary gains in reward coexist with recurrent safety violations.

By comparison, our agent uses fear-triggered expert intervention during early training and requests supervision when the fear signal exceeds a threshold. This improves safety in high-fear encounters and accelerates convergence by providing corrective guidance when the current policy is unreliable. Through these interventions, the agent first learns a reasonably safe and effective policy. With continued training, it internalizes these expert-guided behaviors and gradually refines them into a more efficient and better-calibrated strategy.

Quantitative metrics further support these observations (Fig.~\ref{carla_1}(d)). Time-to-collision increases progressively for SAGE-RL in the later stages of training, while the D3QN baseline declines then partially recovers but remains consistently lower, reflecting SAGE-RL's superior hazard anticipation. Speed profiles show that SAGE-RL transitions from conservative early behavior to confident and stable operation, balancing safety with efficiency. Acceleration variability is also reduced in SAGE-RL, indicating smoother and less aggressive control. Together, these metrics suggest that active expert-guided exploration helps produce policies that better balance safety, efficiency, and smoothness in long-tail hazards.

\begin{figure}[h!]
    \begin{center}
    \includegraphics[width=0.99\linewidth]{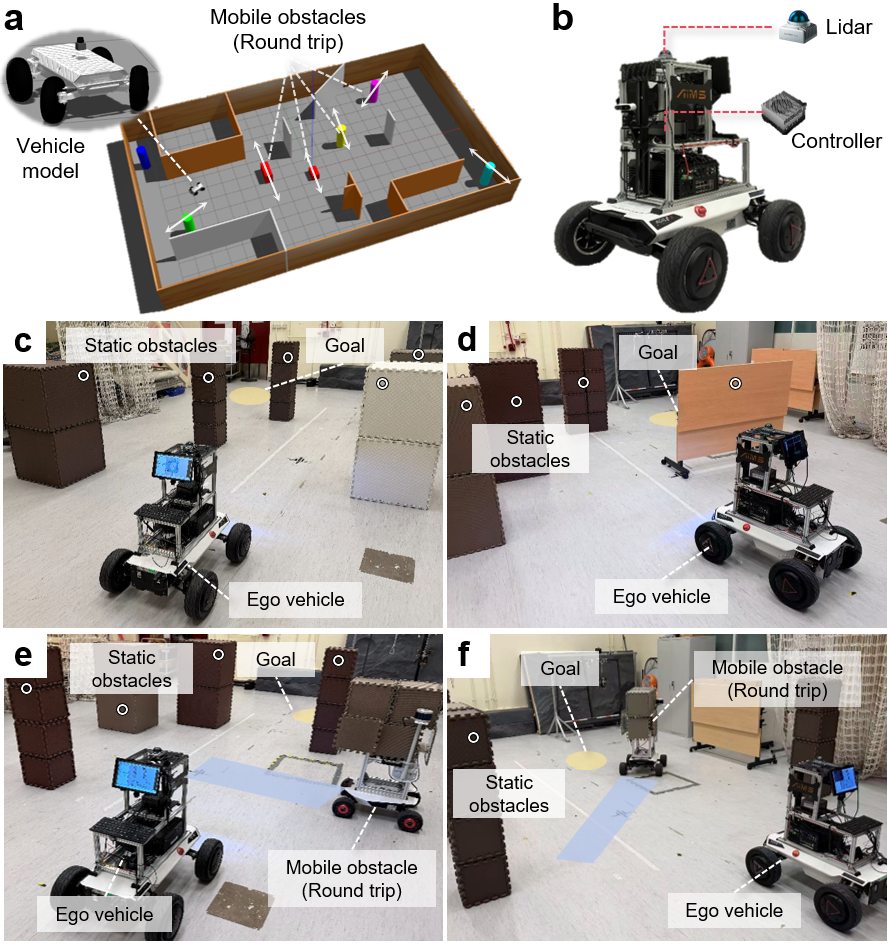}
    \caption{Simulation and real-world validation setup. 
    (a) Gazebo simulation environment matched to the physical UGV platform and obstacle-interaction setting. 
    (b) Real-world UGV platform equipped with LiDAR sensing and onboard computation and control. 
    (c-f) Real-world evaluation scenarios, including static-obstacle tasks and dynamic-obstacle tasks with repeated interactions.} 
    \label{field_test_intro}
 \end{center}                                       
\end{figure}

\subsection{Field Tests}

\subsubsection{Task Description}

Finally, we evaluate the proposed framework on a real unmanned ground vehicle (UGV) to examine continual adaptation under physical execution. To maintain consistency between simulation and hardware experiments, we built the experimental platform and test scenarios shown in Fig. \ref{field_test_intro}. First, a 1:1 replica of the unmanned ground vehicle (UGV) was created in Gazebo (Fig.~\ref{field_test_intro}(a)), matching the map scale, obstacle layout, and the motion pattern of the moving obstacle, so that the simulation conditions closely reflect those in the physical experiments. For real-world tests, we used the UGV platform shown in Fig.~\ref{field_test_intro}(b). The system takes 270° LiDAR point clouds as the primary perception input, while the robot pose is obtained from the LiDAR’s built-in IMU. All perception, planning, and control modules run onboard an NVIDIA Jetson AGX Orin (64 GB) controller to meet real-time computation requirements.

Based on this setup, we design four real-world validation scenarios (Fig.~\ref{field_test_intro}(c-f)): (1) a static-obstacle narrow-passage task, in which the ego vehicle traverses a constrained gap between closely spaced obstacles to reach the goal; (2) a static-obstacle detour task with a large transverse barrier partially blocking the direct route; (3) a dynamic-obstacle task in which a mobile obstacle vehicle repeatedly moves laterally across the ego path; (4) a dynamic-obstacle task involving a frontal approach followed by a diagonal cut-in interaction. In all scenarios, a designated goal region is provided, and the ego vehicle is tasked with navigating from the start to the goal for evaluation.

\begin{table*}[t!]
\centering
\caption{Quantitative comparison of navigation performance across four real-world indoor scenarios.}
\label{tab:all_scenarios_results}
\renewcommand{\arraystretch}{1.2}
\begin{tabular}{l l c c c c c}
\toprule
  & \textbf{Method}
& \makecell{\textbf{Success Rate} \\ (\%) $\uparrow$}
& \makecell{\textbf{Average Speed} \\ (m/s) $\uparrow$}
& \makecell{\textbf{Travel Distance} \\ (m) $\downarrow$}
& \makecell{\textbf{Path Curvature} \\ ($\mathrm{m^{-1}}$) $\downarrow$}
& \makecell{\textbf{Lateral Acceleration} \\ ($\mathrm{m/s^2}$) $\downarrow$} \\
\midrule
\multirow{8}{*}{Scenario 1}
& TEB          & 20 / 20 (100\%) & 0.39 $\pm$ 0.02 & 8.62 $\pm$ 0.19 & 0.489 $\pm$ 0.03 & 0.370 $\pm$ 0.01 \\
& SAC          & 13 / 20 (65\%)  & 0.40 $\pm$ 0.04 & 7.62 $\pm$ 1.46 & 0.560 $\pm$ 0.15 & 0.443 $\pm$ 0.10 \\
& TD3          & 12 / 20 (60\%)  & 0.41 $\pm$ 0.03 & 7.49 $\pm$ 1.48 & 0.493 $\pm$ 0.08 & 0.438 $\pm$ 0.09 \\
& HG-DAgger    & 20 / 20 (100\%) & 0.46 $\pm$ 0.05 & 7.93 $\pm$ 0.20 & 0.487 $\pm$ 0.09 & 0.407 $\pm$ 0.04 \\
& Diffusion-BC & 14 / 20 (70\%)  & 0.40 $\pm$ 0.04 & 7.21 $\pm$ 1.44 & 0.503 $\pm$ 0.04 & 0.405 $\pm$ 0.03 \\
& Diffusion-QL & 16 / 20 (80\%)  & 0.42 $\pm$ 0.04 & 7.34 $\pm$ 0.81 & 0.485 $\pm$ 0.08 & 0.393 $\pm$ 0.06 \\
& Multi-Hug RL & 20 / 20 (100\%) & 0.42 $\pm$ 0.03 & 8.07 $\pm$ 0.22 & 0.505 $\pm$ 0.08 & 0.396 $\pm$ 0.04 \\
& Proposed     & 20 / 20 (100\%) & 0.44 $\pm$ 0.04 & 7.65 $\pm$ 0.13 & 0.472 $\pm$ 0.09 & 0.352 $\pm$ 0.04 \\
\midrule
\multirow{8}{*}{Scenario 2}
& TEB          & 20 / 20 (100\%) & 0.29 $\pm$ 0.02 & 7.52 $\pm$ 0.14 & 0.741 $\pm$ 0.04 & 0.409 $\pm$ 0.02 \\
& SAC          & 12 / 20 (60\%)  & 0.41 $\pm$ 0.04 & 5.16 $\pm$ 1.16 & 0.608 $\pm$ 0.17 & 0.551 $\pm$ 0.13 \\
& TD3          & 8 / 20 (40\%)   & 0.40 $\pm$ 0.08 & 4.92 $\pm$ 1.25 & 0.558 $\pm$ 0.11 & 0.479 $\pm$ 0.08 \\
& HG-DAgger    & 17 / 20 (85\%)  & 0.38 $\pm$ 0.05 & 6.09 $\pm$ 1.03 & 0.662 $\pm$ 0.07 & 0.493 $\pm$ 0.04 \\
& Diffusion-BC & 4 / 20 (20\%)   & 0.33 $\pm$ 0.06 & 4.22 $\pm$ 1.84 & 0.442 $\pm$ 0.08 & 0.428 $\pm$ 0.03 \\
& Diffusion-QL & 10 / 20 (50\%)  & 0.32 $\pm$ 0.04 & 4.97 $\pm$ 1.26 & 0.562 $\pm$ 0.09 & 0.524 $\pm$ 0.06 \\
& Multi-Hug RL & 16 / 20 (80\%)  & 0.36 $\pm$ 0.04 & 5.66 $\pm$ 0.64 & 0.696 $\pm$ 0.08 & 0.592 $\pm$ 0.04 \\
& Proposed     & 20 / 20 (100\%) & 0.35 $\pm$ 0.03 & 5.77 $\pm$ 0.14 & 0.642 $\pm$ 0.10 & 0.508 $\pm$ 0.06 \\
\midrule
\multirow{8}{*}{Scenario 3}
& TEB          & 8 / 20 (40\%)  & 0.28 $\pm$ 0.02 & 5.01 $\pm$ 1.47 & 0.663 $\pm$ 0.17 & 0.369 $\pm$ 0.03 \\
& SAC          & 11 / 20 (55\%) & 0.46 $\pm$ 0.04 & 5.31 $\pm$ 0.67 & 0.664 $\pm$ 0.11 & 0.440 $\pm$ 0.06 \\
& TD3          & 11 / 20 (55\%) & 0.45 $\pm$ 0.04 & 5.03 $\pm$ 1.30 & 0.606 $\pm$ 0.08 & 0.391 $\pm$ 0.05 \\
& HG-DAgger    & 16 / 20 (80\%) & 0.40 $\pm$ 0.05 & 5.80 $\pm$ 0.67 & 0.653 $\pm$ 0.13 & 0.380 $\pm$ 0.03 \\
& Diffusion-BC & 8 / 20 (40\%)  & 0.34 $\pm$ 0.05 & 5.41 $\pm$ 0.89 & 0.588 $\pm$ 0.14 & 0.367 $\pm$ 0.04 \\
& Diffusion-QL & 10 / 20 (50\%) & 0.37 $\pm$ 0.03 & 5.60 $\pm$ 1.34 & 0.633 $\pm$ 0.13 & 0.357 $\pm$ 0.04 \\
& Multi-Hug RL & 15 / 20 (75\%) & 0.44 $\pm$ 0.04 & 6.15 $\pm$ 1.34 & 0.587 $\pm$ 0.05 & 0.380 $\pm$ 0.04 \\
& Proposed     & 20 / 20 (100\%)& 0.43 $\pm$ 0.02 & 6.21 $\pm$ 0.09 & 0.535 $\pm$ 0.04 & 0.356 $\pm$ 0.02 \\
\midrule
\multirow{8}{*}{Scenario 4}
& TEB          & 10 / 20 (50\%) & 0.33 $\pm$ 0.03 & 5.72 $\pm$ 1.50 & 0.625 $\pm$ 0.22 & 0.394 $\pm$ 0.05 \\
& SAC          & 11 / 20 (55\%) & 0.49 $\pm$ 0.06 & 5.02 $\pm$ 0.71 & 0.632 $\pm$ 0.14 & 0.482 $\pm$ 0.06 \\
& TD3          & 9 / 20 (45\%)  & 0.51 $\pm$ 0.03 & 5.06 $\pm$ 0.42 & 0.619 $\pm$ 0.08 & 0.451 $\pm$ 0.09 \\
& HG-DAgger    & 17 / 20 (85\%) & 0.35 $\pm$ 0.02 & 5.58 $\pm$ 0.55 & 0.741 $\pm$ 0.13 & 0.348 $\pm$ 0.04 \\
& Diffusion-BC & 7 / 20 (35\%)  & 0.36 $\pm$ 0.04 & 5.41 $\pm$ 1.01 & 0.677 $\pm$ 0.12 & 0.352 $\pm$ 0.07 \\
& Diffusion-QL & 12 / 20 (60\%) & 0.34 $\pm$ 0.04 & 5.25 $\pm$ 1.14 & 0.669 $\pm$ 0.11 & 0.354 $\pm$ 0.05 \\
& Multi-Hug RL & 16 / 20 (80\%) & 0.38 $\pm$ 0.03 & 6.02 $\pm$ 0.77 & 0.622 $\pm$ 0.11 & 0.421 $\pm$ 0.04 \\
& Proposed     & 20 / 20 (100\%)& 0.46 $\pm$ 0.03 & 6.14 $\pm$ 0.12 & 0.615 $\pm$ 0.11 & 0.367 $\pm$ 0.06 \\
\bottomrule
&& \\[-1em]
\end{tabular}
\begin{minipage}{0.9\linewidth}
  \footnotesize
  \textit{Note:} Average performance is represented by the mean value and the standard deviation.
\end{minipage}
\end{table*}

\subsubsection{Learning Setup}

a) State: The observation comprises a LiDAR encoding and robot-goal state. The 270° front-facing LiDAR field is partitioned into $N=54$ sectors; 
the nearest distance in each sector (clipped and normalized by 10\,m) forms $o_t^{\text{lidar}}\in[0,1]^{54}$. We append normalized goal distance $\bar d_t=\min(d_t/D_{\max},1)$, normalized heading error $\bar\beta_t=\beta_t/\pi$, and previous control commands $(v_{t-1},\omega_{t-1})$. Thus $s_t\in\mathbb{R}^{58}$.

b) Action: Continuous velocity commands $a_t=[v_t,\omega_t]$ (linear velocity and yaw rate) are applied at 10\,Hz.

c) Reward and Cost: The total reward is decomposed into three components:
\begin{equation}
r_t = r_t^{\text{progress}} + r_t^{\text{smooth}} + r_t^{\text{event}} .
\end{equation}
The progress reward encourages the robot to approach the goal,
\(r_t^{\text{progress}} = 10(d_{t-1}-d_t) + 0.5 v_t\).
The smoothness term penalizes aggressive rotational motion,
\(r_t^{\text{smooth}} = -0.5|\omega_t| - \tfrac{1}{5}|\omega_t-\omega_{t-1}|\).
The event-based term accounts for task completion and collisions,
\(r_t^{\text{event}} = -100\,\mathbb{I}_{\text{collision}} + 100\,\mathbb{I}_{\text{goal}}\).

A dense safety cost is defined using the minimum front-laser range $m_t$:
\begin{equation}
c_t = \min\!\left(1,\,
\max\!\left(0,\frac{1.0-m_t}{0.5}\right)\right).
\end{equation}

\subsubsection{Results}

Table~\ref{tab:all_scenarios_results} reports the quantitative comparison of navigation performance across four real-world indoor scenarios. Each method was evaluated over 20 trials per scenario using success rate, average speed, travel distance, path curvature, and lateral acceleration as performance metrics. Overall, the proposed method achieves a 100\% success rate in all four scenarios, demonstrating more reliable goal-reaching performance than the learning-based baselines and maintaining robustness across different indoor layouts. Compared with SAC, TD3, Diffusion-BC, Diffusion-QL, and Multi-Hug RL, our method consistently improves navigation reliability, especially in challenging scenarios where several baselines suffer from degraded success rates. Although TEB also achieves perfect success in Scenarios 1 and 2, its performance drops in Scenarios 3 and 4, indicating limited adaptability in more complex real-world conditions. In addition to reliability, the proposed method maintains competitive motion quality, with relatively low path curvature and lateral acceleration, suggesting smoother and more stable navigation behavior while keeping speed and path length within a competitive range among successful methods.

\begin{figure*}
    \begin{center}
    \includegraphics[width=0.99\linewidth]{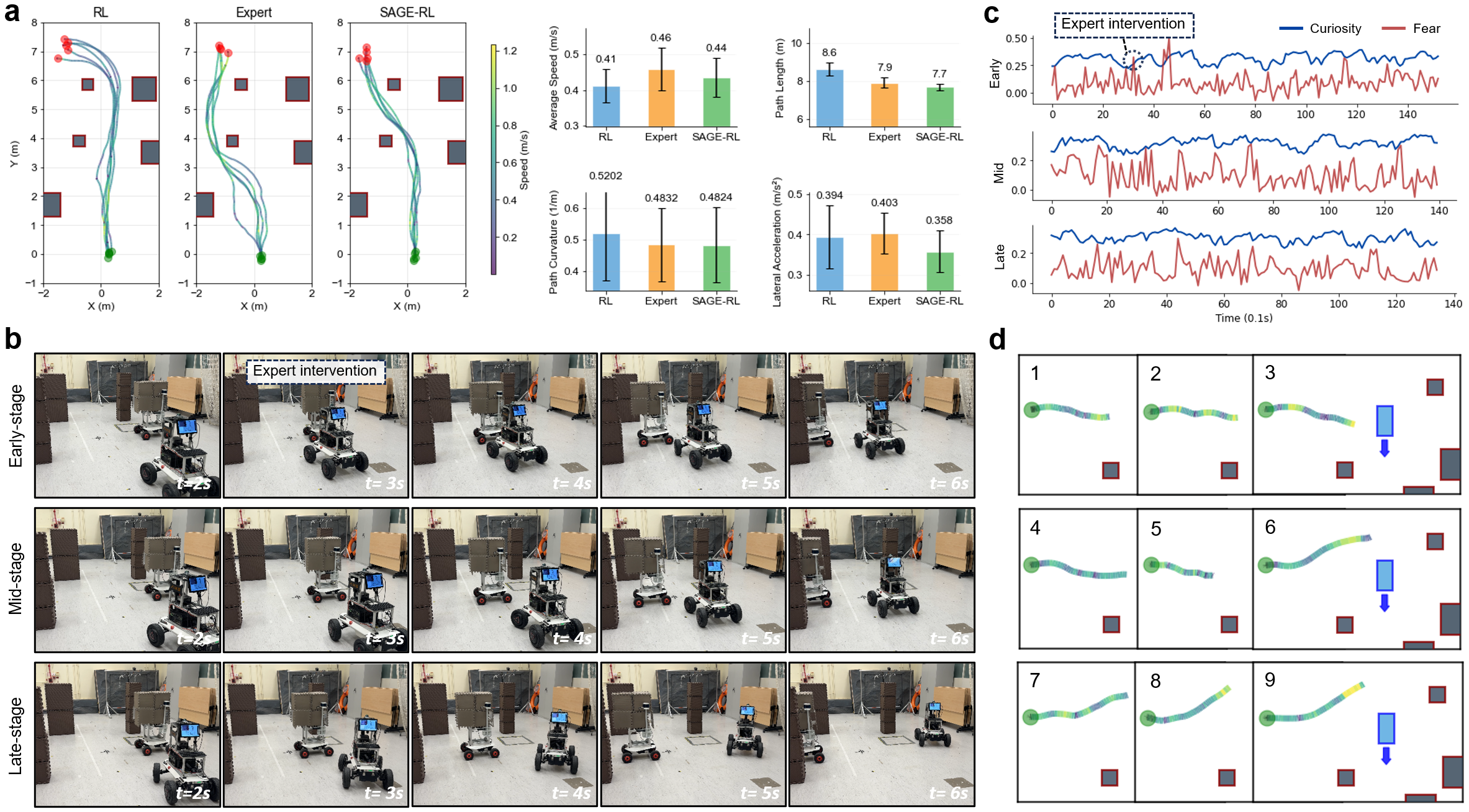}
    \caption{Continual adaptation in real-world navigation tests. Scenario 1:
    (a) Representative trajectories and quantitative comparison in the static scenario.
    Scenario 3:
    (b) Snapshots from early, middle, and late stages of continual fine-tuning in the dynamic scenario, showing behavioral refinement through repeated interaction and expert intervention when necessary; 
    (c) Fear signal and curiosity-regulated intervention threshold during deployment, indicating the timing of expert takeover;
    (d) Trajectories over nine consecutive deployment-fine-tuning cycles in the dynamic scenario, showing progressively refined behavior.} 
    \label{field_test_1}
 \end{center}                                       
\end{figure*}

Fig. \ref{field_test_1} summarizes the real-world evaluation in scenarios 1 and 3, and highlights the real-world continual adaptation enabled by our method. In the static scenario 1 (Fig.~\ref{field_test_1}(a)), we compare representative trajectories produced by different methods and report the corresponding quantitative metrics. The results show that our approach reliably reaches the goal while maintaining competitive path quality and motion smoothness under fixed obstacle layouts.

More importantly, in the dynamic scenario 3 (Fig.~\ref{field_test_1}(b-d)), we show how the proposed system improves through repeated trials of the same task with continual fine-tuning. Fig.~\ref{field_test_1}(b) presents snapshots from the early, middle, and late stages of this process. In the early stage, the vehicle may produce suboptimal or potentially unsafe responses when interacting with the moving obstacle, causing the fear signal to exceed the intervention threshold and trigger expert takeover. Through these interventions, the agent selectively imitates expert behavior to avoid collision and gradually learns a safer policy. With further fine-tuning, it handles the same interaction more smoothly and autonomously. In some trials, the adapted policy follows a shorter detouring trajectory than the conservative expert response. Consistent with these observations, Fig.~\ref{field_test_1}(c) shows the evolution of the fear signal, the curiosity-regulated intervention threshold, and the interval of expert takeover. 

Providing a trajectory-level view of the same process, Fig.~\ref{field_test_1}(d) summarizes behavior over nine consecutive deployment-fine-tuning cycles and shows progressively refined motion in the dynamic scenario 3. Across these cycles, the trajectories are progressively refined. In particular, when the lateral obstacle cuts in, the expert typically responds by stopping abruptly to avoid collision. After continual fine-tuning, however, the agent gradually improves beyond this strategy and discovers a path that detours behind the obstacle vehicle, thereby reducing unnecessary stopping. Overall, these field tests show that the proposed framework supports continual adaptation in real-world navigation through active learning and expert-supervised updates, improving behavior beyond the initial policy.

\section{Conclusion}\label{Sec_Conclusion}

This work addresses a key challenge in learning-based AD: policies that perform well on the dominant training distribution may still fail abruptly in rare, safety-critical situations. To mitigate this issue, we propose a self-awareness-guided active learning framework for continual adaptation. The agent uses a short-horizon fear signal as a proxy for prospective risk, while curiosity modulates the intervention threshold. When fear exceeds this adaptive threshold, expert takeover limits unsafe exploration and converts the interaction into targeted supervision. Across route-level distribution shifts, adversarial traffic interactions, occluded pedestrian hazards, and real-world UGV navigation tasks, we observe a consistent pattern: elevated fear triggers intervention in high-risk states, policy behavior stabilizes with experience, and intervention becomes less frequent as the agent adapts.

The main contribution is a closed-loop mechanism that links self-monitoring, active expert intervention, and targeted policy improvement. Rather than relying solely on stronger policy architectures or more offline data, the proposed framework uses internal risk estimates to decide when autonomous exploration should be guided. This concentrates expert supervision on safety-critical and information-rich moments, improving safety during learning and increasing the relevance of collected experience for policy updates. Since fear estimation and expert intervention operate at the interaction and data-collection level, the framework can in principle be integrated with other data-driven policy-learning methods and extended beyond driving to other embodied safety-critical systems.

Several limitations remain. The fear signal may be miscalibrated under severe perception degradation, sensor corruption, or highly atypical scenarios, leading to delayed or unnecessary interventions. Expert takeover also introduces supervision cost and may make early training more conservative. Thus, the framework should be viewed not as a guarantee of absolute safety, but as a practical approach for reducing risk during continual adaptation. Future work will focus on improving fear calibration and interpretability, designing more data-efficient intervention schedules, and extending the framework to longer-horizon online adaptation with stronger safety assurance.


\bibliographystyle{IEEEtran}

\small\bibliography{Bibliography}
\end{document}